\documentclass[letterpaper, 10 pt, conference]{ieeeconf}  
\IEEEoverridecommandlockouts                              

\usepackage{flushend}
\usepackage{lipsum}
\usepackage[utf8]{inputenc}
\usepackage{graphicx}
\usepackage{color}
\usepackage{latexsym}
\usepackage{bmpsize}
\usepackage{xurl}
\usepackage{comment}
\usepackage{amsmath,amssymb}
\usepackage[nobreak]{cite}
\usepackage[subrefformat=parens]{subcaption}
\usepackage[hyperfootnotes=false]{hyperref}
\usepackage{cleveref} 
\usepackage{here}
\usepackage{listings} 
\usepackage[affil-it]{authblk} 

\crefname{figure}{Fig.}{Fig.}
\crefname{table}{Table}{Table}
\crefname{section}{Section}{Section}
\newcommand{\todo}[1]{}
\renewcommand{\todo}[1]{{\color{red} {\bf TODO:#1}}}
\newcommand{\circletext}[1]{\raise0.2ex\hbox{\textcircled{\scriptsize{#1}}}}

\def\Underline{\setbox0\hbox\bgroup\let\\\endUnderline}
\def\endUnderline{\vphantom{y}\egroup\smash{\underline{\box0}}\\}

\definecolor{myComment}{rgb}{0.0, 0.6, 0.0}       
\definecolor{myKeyWord}{cmyk}{1.0, 0.0, 0.0, 0.3} 
\definecolor{myString}{cmyk}{0.0, 1.0, 0.0, 0.0}  
 
\lstdefinestyle{customText}{
    backgroundcolor  = {\color{white}},               
    basicstyle       = {\footnotesize},               
    breaklines       = {true},                        
    commentstyle     = {\itshape  \color{myComment}}, 
    keywordstyle     = {\bfseries \color{myKeyWord}}, 
    lineskip         = {-0.5ex},                      
    showstringspaces = {false},                       
    sensitive        = {true},                        
    stepnumber       = {1},                           
    stringstyle      = {\ttfamily \color{myString}},  
    tabsize          = {2},                           
    xleftmargin      = {2zw},                         
    xrightmargin     = {2zw}                          
}
\begin{document}

\title{\bf{\LARGE{{Realtime Trajectory Smoothing with Neural Nets}}}}

\author[1,2]{Shohei Fujii\thanks{E-mail: \href{mailto:SHOHEI001@e.ntu.edu.sg}{SHOHEI001@e.ntu.edu.sg}; Corresponding author} }
\author[1,3]{Quang-Cuong Pham}
\affil[1]{School of Mechanical and Aerospace Engineering, Nanyang Technological University, Singapore}
\affil[2]{DENSO CORP., Japan}
\affil[3]{Eureka Robotics, Singapore}
\maketitle

\begin{abstract}
  In order to safely and efficiently collaborate with humans,
  industrial robots need the ability to alter their motions quickly to
  react to sudden changes in the environment, such as an obstacle
  appearing across a planned trajectory. In Realtime Motion Planning,
  obstacles are detected in real time through a vision system, and new
  trajectories are planned with respect to the current positions of
  the obstacles, and immediately executed on the robot. Existing
  realtime motion planners, however, lack the smoothing
  post-processing step -- which are crucial in sampling-based motion
  planning -- resulting in the planned trajectories being jerky, and
  therefore inefficient and less human-friendly. Here we propose a
  Realtime Trajectory Smoother based on the shortcutting technique to
  address this issue. Leveraging fast clearance inference by a novel
  neural network, the proposed method is able to consistently smooth
  the trajectories of a 6-DOF industrial robot arm within 200 ms on a
  commercial GPU. We integrate the proposed smoother into a full
  Vision--Motion Planning--Execution loop and demonstrate a realtime,
  smooth, performance of an industrial robot subject to dynamic
  obstacles.
  
\end{abstract}

\section{Introduction}

In order to safely and efficiently collaborate with humans, robots
need the ability to alter their motions quickly to react to sudden
changes in the environment, such as an obstacle appearing across a
planned trajectory. In most industrial applications, one would stop
the robot upon the detection of obstacles in the robot's reach
space. However, such a solution is inefficient and precludes true
human-robot collaboration, where humans and robots are to share a
common workspace.

Recently, Realtime Motion Planning (RMP) has been proposed to enable
true human-robot collaboration: obstacles are detected in real time
through a vision system, new trajectories are planned with respect to
the current positions of the obstacles, and immediately executed on
the robot. RMP requires extremely fast computation in the
Vision--Motion Planning--Execution loop. In particular, several
techniques have been proposed for the Motion Planning component,
relying on the parallelization of sampling-based
algorithms~\cite{RRT,PRM} on dedicated hardware, such as
GPU~\cite{jiaGPlannerRealTimeMotion, panGPUbasedParallelCollision2010}
or FPGA~\cite{murrayRobotMotionPlanning2016}.

Sampling-based motion planners typically output jerky trajectories
and therefore almost always require a smoothing post-processing
step~(see e.g. \cite{luo2014empirical} for a detailed review). To our
knowledge, existing realtime motion planners lack this
step\,\footnote{Video:
  Yaskawa Motoman Demo by Realtime Robotics on Vimeo
  \\ \url{https://vimeo.com/359773568}.  Note the jerky transitions,
  for example at time stamps 24s, 30s, and 35s.}, presumably because
of the large computation time associated with trajectory
smoothing. Indeed, to obtain an acceptable trajectory quality,
\emph{smoothing time is comparable, if not longer than initial path
  planning time}~\cite{luo2014empirical}. As a result, while the
motions produced by Realtime Motion Planning enable safely adapting
to sudden changes in the environment, the lack of smoothness makes
them inefficient and less human-friendly. In the particular context of
human-robot collaboration, smooth trajectories indeed appear more
predictable and agreeable to humans (see e.g.~\cite{pham2015new} for a
discussion).

\begin{figure}[t]
  \begin{center}
    \includegraphics[width=0.9\linewidth,trim={0cm 0cm 0cm 1cm},clip]{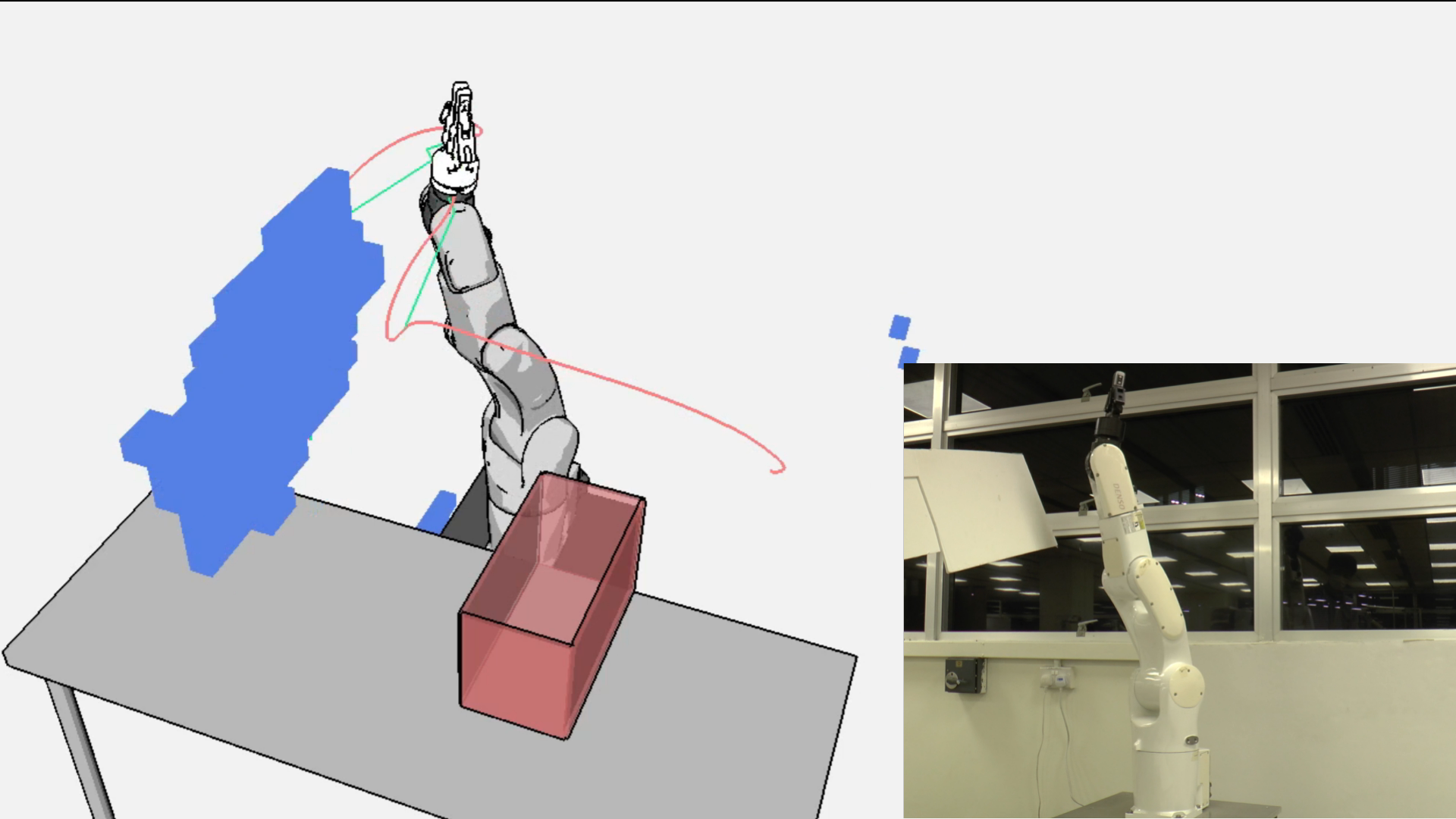} 
    \caption{A robot avoids a dynamic obstacle by realtime trajectory
      re-planning and smoothing with the proposed smoother. Green:
      jerky trajectory planned by realtime PRM. Red: trajectory after
      smoothing. See the video of the experiment at
      ~\url{https://youtu.be/XQFEmFyUaj8}.
      }
  \label{fig:replanning_snapshot}
  \end{center}
\end{figure}

Here we propose a Realtime Trajectory Smoother based on the
shortcutting
technique~\cite{geraerts2007creating,hauserFastSmoothingManipulator2010}
to address this issue. Leveraging fast clearance inference by a novel
neural network, the proposed method is able to consistently smooth the
trajectories of a 6-DOF industrial robot arm within 200 ms on a
commercial GPU, which is 2 to 3 times faster than state-of-the-art
smoothers. Combined with even rudimentary, in-house, implementations
of a vision pipeline and a sampling-based motion planner, we were able
to achieve 300 ms cycle time, which is sufficient for realtime
performance (\cref{fig:replanning_snapshot}). Note again here that
smoothing is indeed the bottleneck, as the initial planning time was
only $\sim 40$ ms.

The paper is organized as follows.
We survey related work on obstacle avoidance and collision estimation in \cref{sec:related_work}.
In \cref{sec:our_method}, we
present our trajectory smoothing pipeline and the structure of the
neural network model for fast clearance inference. In
\cref{sec:experiment}, we evaluate the pipeline in two sets of
experiments. First, we evaluate the clearance inference accuracy of
the neural network. Second, we integrate the smoother into a
full-fledged Vision--Planning--Execution loop, and demonstrate a
realtime, smooth, performance of a physical industrial robot subject
to dynamic obstacles. Finally, we discuss the advantages and
limitations of our approach and conclude with some directions for future
work~(\cref{sec:conclusion}).

\section{Related Work} \label{sec:related_work}
In this section, we give a brief overview of prior work on realtime
motion planning in dynamic environments, trajectory smoothing,
optimization-based planning and reactive control.

\subsection{Realtime Motion Planning}
The major sampling-based planners are Rapidly-exploring Random Tree
(RRT)\cite{RRT} and Probabilistic Roadmaps (PRM)\cite{PRM}, and both
have realtime versions.  RT-RRT* is a tree rewiring technique for RRT
which keeps removing nodes and edges colliding with dynamic obstacles
and adding new nodes and edges\cite{naderiRTRRTRealtimePath2015}.
g-Planner is an RRT-based technique to utilize GPU's parallel
computation mechanism to accelerate configuration node sampling and
collision checking\cite{jiaGPlannerRealTimeMotion,
  panGPUbasedParallelCollision2010}.  Parallel Poisson RRT also
exploits GPU for tree expansion, nearest neighbor search, and collision
checking\cite{parkParallelMotionPlanning2017}.  On the other hand,
Murray et al.\ \cite{murrayRobotMotionPlanning2016} use FPGA for a
PRM-based planner by voxelizing the environment and caching collision
information into each edge of PRM's roadmap.  Although these methods
successfully accelerate sampling-based motion planning, they still
require fast smoother of their generated path, to apply their
technologies onto a real robot.

\subsection{Trajectory Smoothing}

The idea of path smoothing by shortcutting was first proposed by
Geraerts and Overmars in~\cite{geraerts2007creating}.  Hauser et
al. extend the method by introducing a parabolic trajectory
representation, enabling taking into account velocity and acceleration
bounds~\cite{hauserFastSmoothingManipulator2010}. Later, Ran et al.
extend this parabolic smoothing algorithm into cubic smoothing
algorithm with jerk
constraints\cite{zhaoTrajectorySmoothingUsing2015}.  Besides, Pan
et al. introduced b-spline based trajectory representation and its
smooth shortcutting
algorithm\cite{panCollisionfreeSmoothTrajectory2012}.  All of these
methods can successfully compute a smooth path from piecewise linear
trajectory, but they are computationally slow because of many
queries on collision checking,

\subsection{Optimization-based Motion Planning}
Another approach is optimization-based trajectory generation.  The
first optimization-based approach is
CHOMP\cite{ratliffCHOMPGradientOptimization2009}, where they formulate
motion planning as a quadratic problem and solve it as a sequential
optimization problem by iterative linearization.  To remove dependency
on the computation of gradients, STOMP was introduced by Kalakrishnan
et al.\ \cite{kalakrishnanSTOMPStochasticTrajectory2011}.  Although
these methods can successfully compute a smooth trajectory, they are
intrinsically slow because optimization needs to go down
following the gradient of a cost function.  Real-time optimization-based
planner using GPU was proposed in
\cite{parkRealtimeOptimizationbasedPlanning2013}, showing its
capability to avoid dynamic obstacles by introducing parallel
threading optimization from different random seeds.  However, it is
uncertain that this method works well as well in an environment with
many obstacles due to its dependency on randomness.  An approach to
integrate classic sampling-based planner with optimization-based planner 
\cite{daiImprovingTrajectoryOptimization2018} cannot still handle dynamic obstacles.

\subsection{Reactive Control}
Another approach is to control robots reactively according to the
change in the environment.  Kapper et al.\ introduce reactive motion
generation which applies a virtual power to the robot and leads it to
avoid obstacles locally\cite{kapplerReactiveMotionGeneration2018}.
Relaxed IK is a technique to compute inverse kinetics while
considering the continuity of the solution to the current robot joint
values\cite{relaxedik_Rakita-RSS-18}.  Although these approaches can
generate a smooth motion, it has a possibility to get stuck into local
minima and not to get out of it.

\subsection{Collision Estimation by Machine Learning}
There are several papers which estimate collisions/clearances from a
robot to its environment using machine learning techniques such as
CN-RRT \cite{kewNeuralCollisionClearance2019} and Fastron
series\cite{dasFastronOnlineLearningbased2017,
  zhiDiffCoAutoDifferentiableProxy2021}.  Although they successfully
adapt clearance estimation into motion planning (CN-RRT) and
optimization (DiffCo), CN-RRT cannot handle dynamic obstacles fast and
adaptively, and DiffCo's optimization-based smoothing takes more than
2 seconds, whereas our method can smooth a trajectory within 0.2--0.3
seconds while handling dynamic obstacles without any runtime
modification onto a neural network or any other parameters including
geometric collision checking for safety.

\section{Realtime Trajectory Smoothing}\label{sec:our_method}
\subsection{Overview}

\begin{figure*}[thbp]
  \begin{center}
  \includegraphics[width=1.0\textwidth]{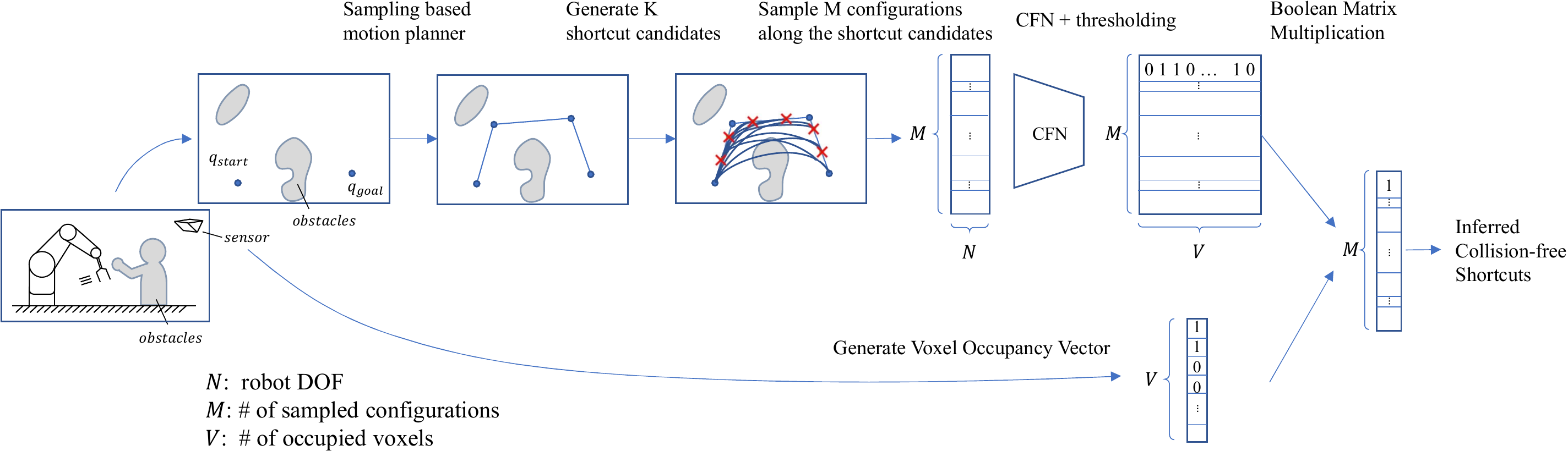}
  \caption{A pipeline of trajectory collision estimation in NN-accelerated trajectory smoother. See \cref{sec:our_method} for detail.}
  \label{fig:method_detail}
  \end{center}
  \vspace{-2mm}
\end{figure*}

The pipeline of our NN-based trajectory smoother is illustrated in
\cref{fig:method_detail}. Consider a $N$-DOF robot. Given a piecewise
linear path (blue lines) in the configuration space, typically
outputted by a sampling-based planner, our smoother samples $c$
configurations at regular time-intervals (red X marks), and computes
parabolic shortcut trajectories between every pair of sampled
configurations.
We have $(c+2)$ configurations in total on a given path which consists of 
the $c$ sampled waypoints plus a starting configuration $q_{start}$ and a goal configuration $q_{goal}$.
Let them be $q_0 ... q_{c+1}$,
and a shortcut from $q_i$ to $q_j$ be $S(q_i, q_j)$. We have
$K=\frac{(c+2)(c+1)}{2}$ shortcut candidates, namely, $S(q_0, q_1)$,
$S(q_0,q_2)$, ... $S(q_0,q_{c+1})$, $S(q_1,q_2)$, $S(q_1, q_3)$
... $S(q_1, q_{c+1})$, ... $S(q_c,q_{c+1})$.

Given a shortcut $S(q_i, q_j)$, we sample $m_{ij}$ configurations at
regular intervals along the shortcut. Next, we stack the
$M=\sum_{ij}m_{ij}$ configurations into one single $M \times N$
matrix. This matrix is then fed into the ``Clearance Field
Neural Network'' (CFN, see details in \cref{subsec:CFN}) for
\emph{batch processing}. Assume that the spatial workspace is
discretized into $V$ voxels, the CFN returns a matrix of size
$M \times V$ containing the inferred clearances from the robot placed
at every sampled configuration to every voxel of the discretized
spatial workspace. Next, we perform thresholding to obtain the
$M \times V$ Inferred Collision Matrix: a configuration $q$ is
considered as in collision with a voxel if the clearance from the
robot placed at $q$ to the voxel is smaller than a given threshold.

In parallel, from the realtime pointcloud captured by the vision
system, we generate the $V\times 1$ Voxel Occupancy Vector: an element
of this vector is 1 if the corresponding voxel is occupied by an
obstacle, 0 if not.

Next, we perform a boolean matrix multiplication of the $M\times V$
Inferred Collision Matrix by the $V\times 1$ Voxel Occupancy Vector to
obtain a $M \times 1$ vector that gives the inferred collision status
for all the sampled configurations, which in turn yields the inferred
collision status for all the $K$ shortcut candidates (a shortcut
candidate is in collision if any of its sampled configurations is in
collision).

Finally, we run the Dijkstra algorithm to find the shortest trajectory
consisting of \emph{inferred collision-free} shortcuts. We then check
the \emph{actual} collision status of the obtained shortcutted
trajectory by a geometric collision checker. If the inference is
exact, the shortcutted trajectory should be collision-free and
selected. If not, we re-run Dijkstra until an actually collision-free
trajectory is found.  In practice, owing to the good inference
quality, we observed that the shortcutted trajectory returned by the
first Dijkstra call is actually collision-free 86.1\% of the time.

\subsection{Clearance Field Network (CFN)}
\label{subsec:CFN}

\begin{figure}[t]
  \begin{center}
  \vspace{-2mm}
  \includegraphics[width=0.85\linewidth,trim={4cm 3cm 1cm 6cm},clip]{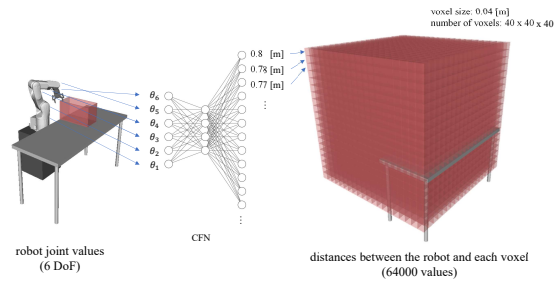}
  \caption{The environment is discretized into $V$ voxels. A voxel's
    clearance is the clearance between the voxel and the robot
    surface. A Clearance Field is a $V\times 1$ vector that contains
    all the clearances of the $V$ voxels. As the Clearance Field
    depends on the robot configuration, a Clearance Field Network
    learns the mapping from a configuration $q$ to its
    ClearanceField($q$).}
  \label{fig:voxelwise_clearances_experiment_settings}
  \vspace{-4mm}
  \end{center}
\end{figure}

We formally define the clearance of a voxel as the signed-distance
between the voxel and the robot surface. Note that the clearance can
be negative if the voxel is ``inside'' the robot. The Clearance Field
is then defined as a $V\times 1$ vector that contains all the
clearances of the $V$ voxels. Observe that the Clearance Field depends
on the robot configuration. A Clearance Field Network is a Neural
Network that learns the mapping
(see \cref{fig:voxelwise_clearances_experiment_settings} as well):
\begin{eqnarray}
  \mathbb{R}^N &\rightarrow& \mathbb{R}^V \nonumber\\
  q & \mapsto & \mathrm{ClearanceField}(q). \nonumber
\end{eqnarray}

To learn this mapping, we follow a supervised learning approach:
offline, we generate a large number of random configurations. For each
configuration $q$, we use a geometric collision-checker (which
provides clearance data, such as FCL~\cite{FCL}) to calculate the
clearance at every voxel, constructing thereby ClearanceField($q$). At
run time, given a new, possibly unseen $q_\mathrm{new}$, one can
quickly infer ClearanceField($q_\mathrm{new}$).

\begin{figure}[tbhp]
  \begin{center}
  \includegraphics[width=0.9\linewidth]{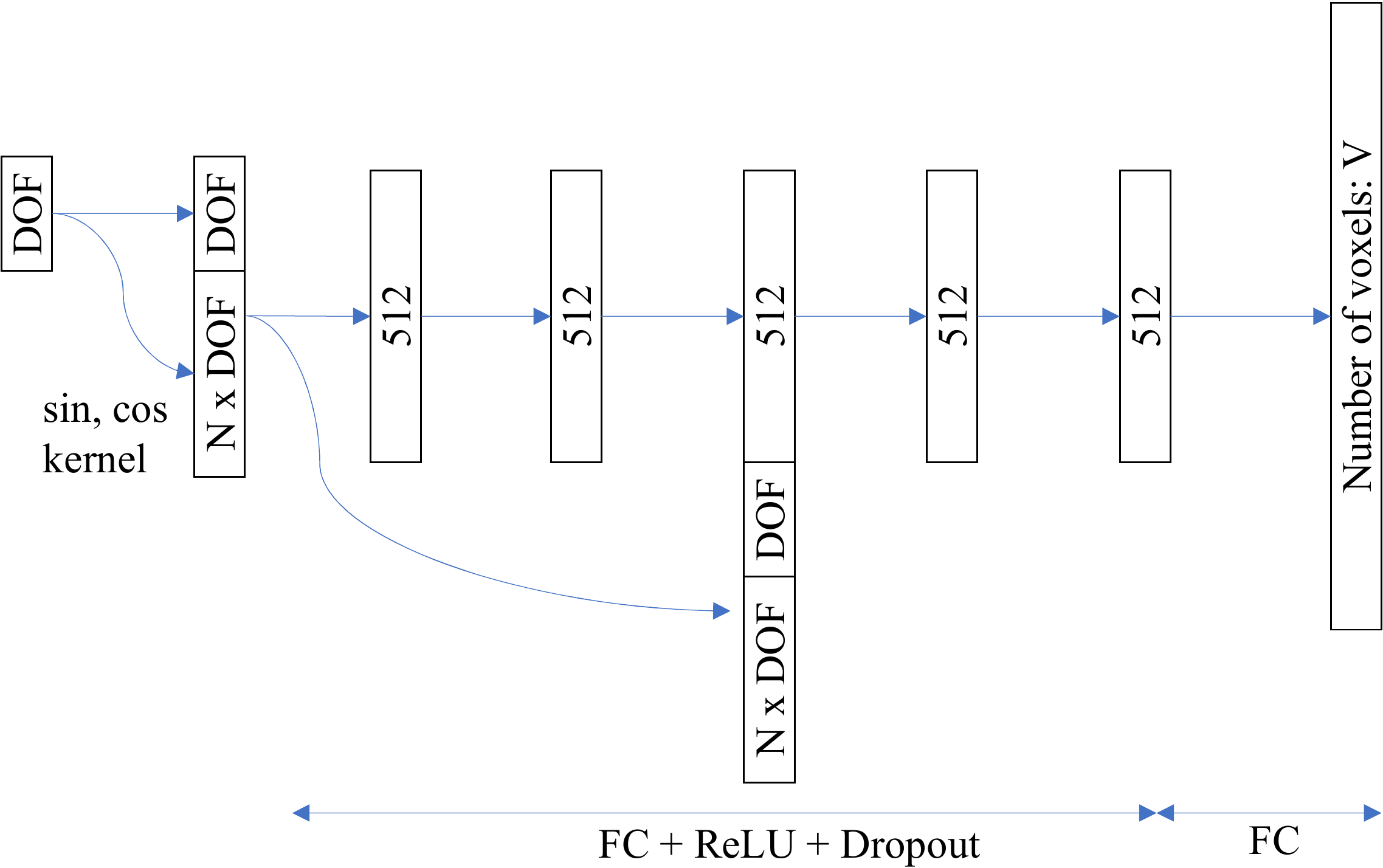}
  \caption{The architecture of Clearance Field Network: Kernel
    function inputs a high-frequency values into a neural network
    using sine and cosine. The middle layers are composed of
    Fully-Connected, ReLU and DropOut with one skip connection.}
  \label{fig:network_architecture}
  \vspace{-2mm}
  \end{center}
\end{figure}

The architecture of the proposed Clearance Field Network is shown in
\cref{fig:network_architecture}.  The ``sin, cos kernel'' converts
joint values $q$ to: 
\begin{equation}
  \begin{aligned}
    \textrm{ker}(q) = [\sin(2^0 \pi q), & \cos(2^0 \pi q), ..., \\
    &\sin(2^{L-1} \pi q), \cos(2^{L-1} \pi q)]
  \end{aligned}
\end{equation}
inspired by NeRF's positional encoding \cite{mildenhall2020nerf}.
This is intended to increase the frequency of input values and allow
the neural network only to learn low-frequency features.

The advantage of this method is that it can handle dynamic
obstacles. Previously, \cite{kewNeuralCollisionClearance2019} need to
feed the information of dynamic obstacles into the neural network.
However, since the dimension of input of the neural network should be
static and fixed, it is needed to convert the variable size of
information about dynamic obstacles into a static-sized feature
vector, and feed it into the neural network. In contrast, in our
approach, the dynamic obstacles already exist in the form of occupied
voxels, whose number is fixed. Our approach can thus apply to any
number/shape of dynamic obstacles and the computation can be easily
parallelized on GPU.

The reason why we propose to learn clearances instead of directly
collision status is that neural networks are better at
approximating continuous functions, and clearance is a
Lipschitz-continuous function of the
environment\cite{kewNeuralCollisionClearance2019}, while collision
status is a discrete function.

\section{Experiments and Results} \label{sec:experiment}

\subsection{Performance of CFN}
First, we examine the clearance field network.  We train our neural
network with 52,000 joint values and their corresponding clearances
using a batch size of 50, validate the training process with 16,000
validation data, and test the trained network with 12,000 test
data. During training, we use L1 loss and Adam optimizer
\cite{kingmaAdamMethodStochastic2015} with a learning rate of
$1 \times 10^{-3}$. 
We set $L = 3$ for positional encoding in this experiment.  
The total data generation takes 2.7 days using 16
CPU cores.  The optimization takes 300 iterations (about 1 hour) to
converge. Note that the above data generation and optimization steps
need to be done only once for a given robot model (without
obstacles).
The obstacles are handled by the fast \emph{inference} step at execution time.


\begin{figure}[tbhp]
  \begin{center}
  \begin{minipage}[b]{1.0\hsize}
    \includegraphics[width=0.49\linewidth,trim={0cm 0cm 1cm 1cm},clip]{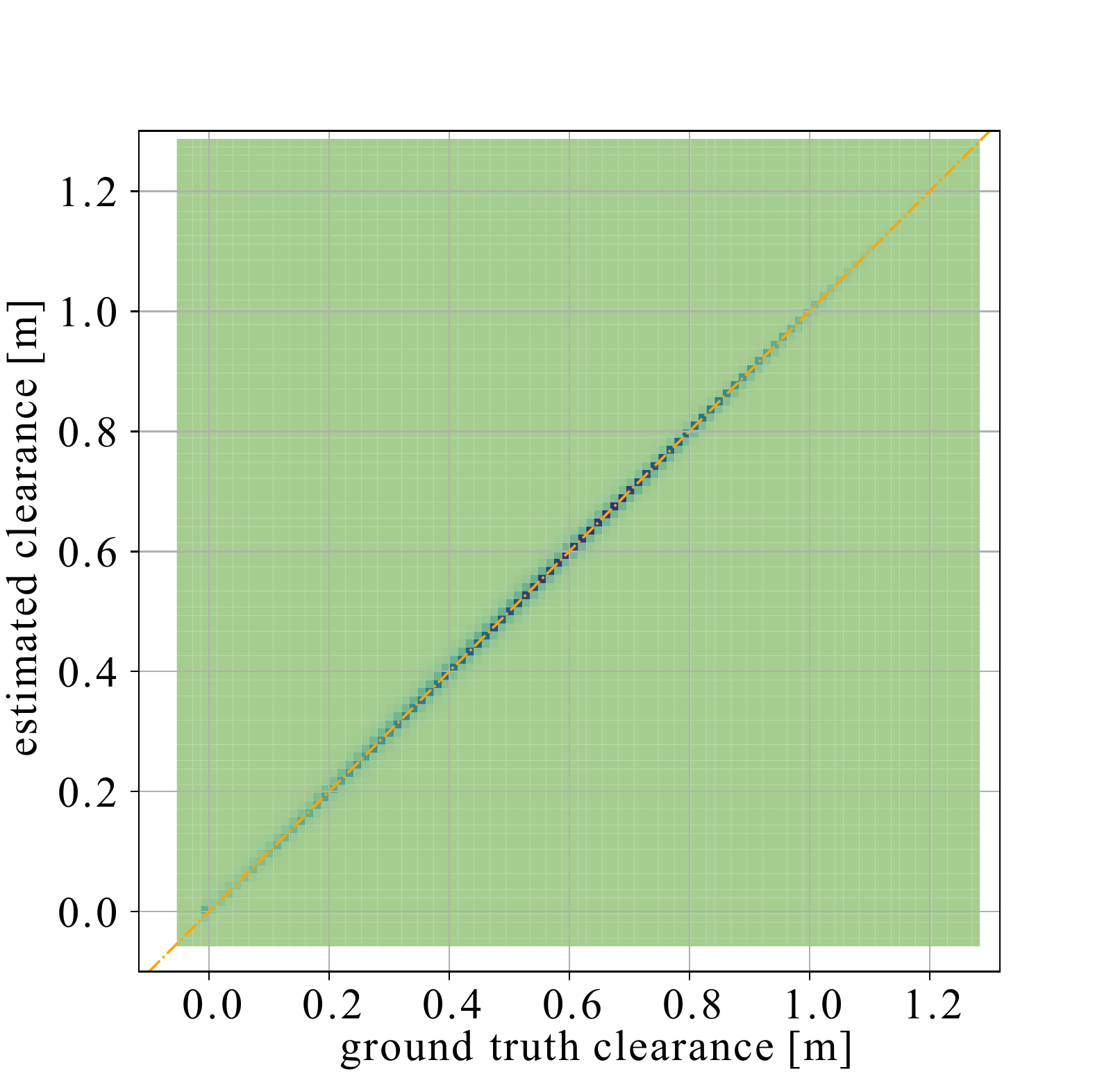}
    \includegraphics[width=0.49\linewidth,trim={2cm 0cm 0cm 2cm},clip]{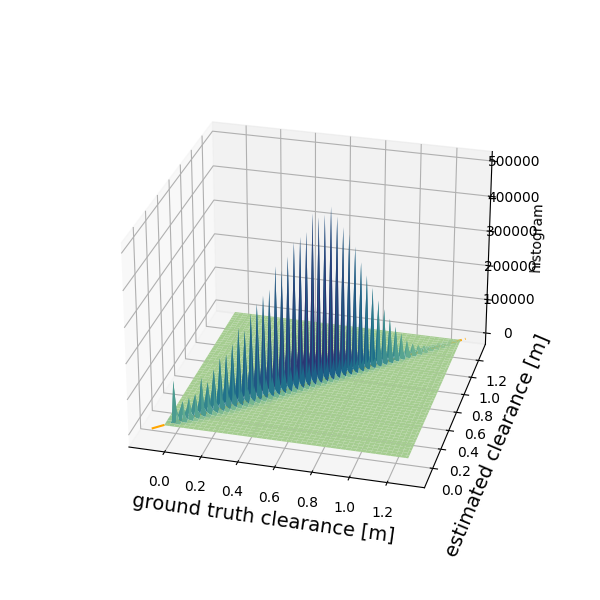}
    \subcaption{All test results}
    \label{fig:clearance_error_histogram_overview}
  \end{minipage}
  \begin{minipage}[b]{1.0\hsize}
    \includegraphics[width=0.49\linewidth,trim={0cm 0cm 0cm 1cm},clip]{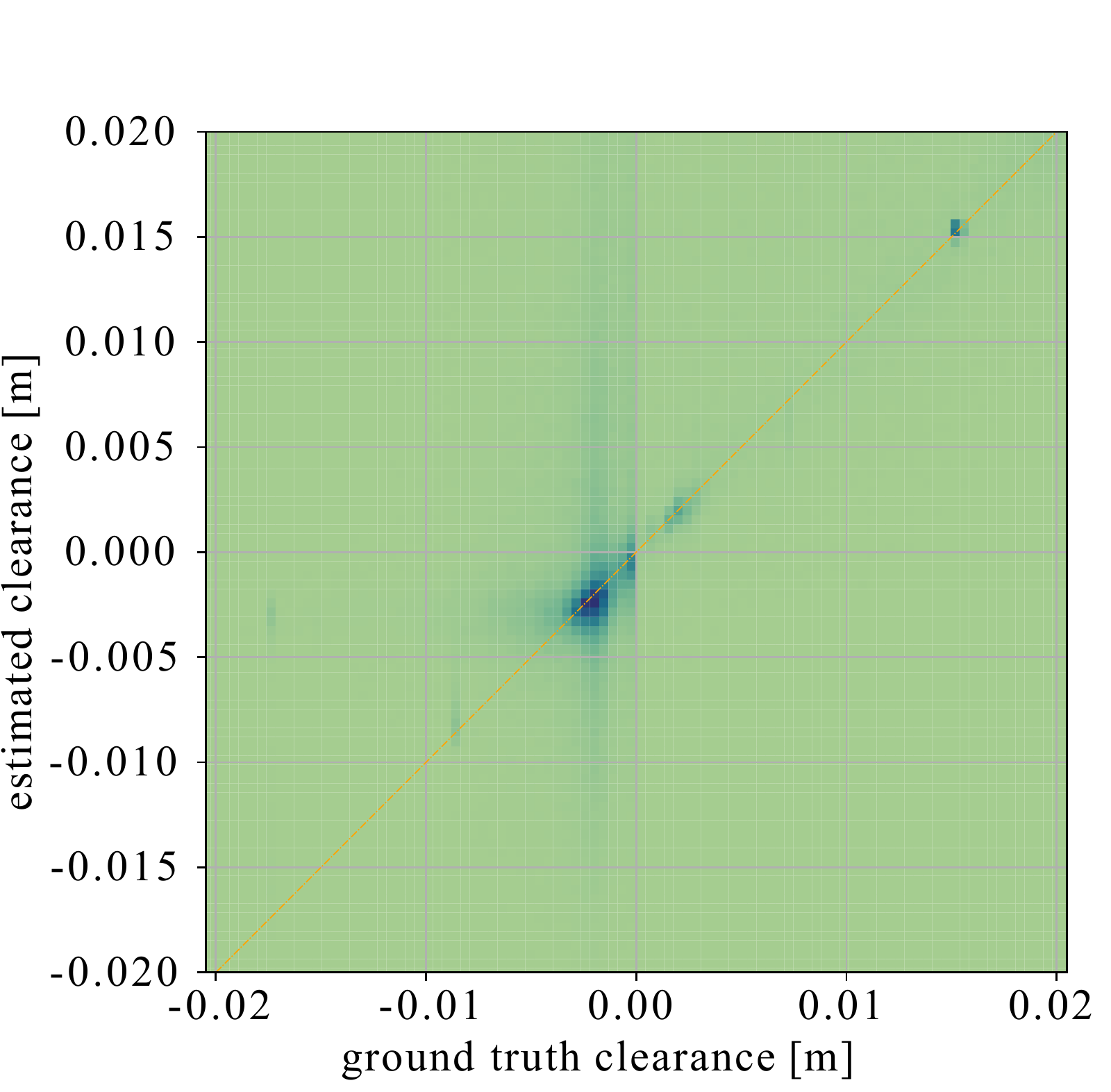}
    \includegraphics[width=0.49\linewidth,trim={2cm 0.8cm 0cm 2.2cm},clip]{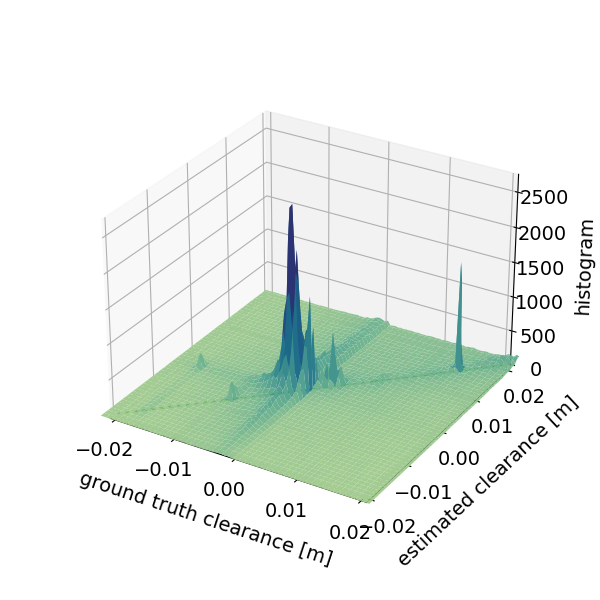}
    \subcaption{Zoomed in the origin}
    \label{fig:clearance_error_histogram_overview_zoomed}
  \end{minipage}
  \caption{Colored histograms of Clearance Field Network estimation.
  Upper: a 2D histogram using all 12,000 test data (left) and its 3D mesh plotting (right).
  Lower: zoomed in the origin with higher resolution (left) and its 3D mesh plotting (right). An orange dotted line represents a 45 degree line for a reference.
  }
  \label{fig:clearance_estimation_graphs}
  \vspace{-4mm}
  \end{center}
\end{figure}

We show colored 2D/3D histograms of clearances estimation in \cref{fig:clearance_estimation_graphs}.
False negatives (i.e. robot position in collision being classified as free) lead to invalid trajectory and more time for shortcutting, whereas false positives (i.e. collision-free robot position being classified as in collision) lead to conservative shortcutting and a longer trajectory.
This trade-off 
can be managed by a clearance threshold.
Precision ($\frac{\mathrm {Estimated\ as\ in\-collision}}{\mathrm {Actually\ in\-collision}}$) is 85.3\% and 90.9\% when we set a threshold as 20 [mm] and 30 [mm] respectively, which are sufficient to infer the collision status of trajectories.
By setting a large threshold, we can reduce the probability to obtain a geometrically in-collision shortcutted trajectory while it may lose a shorter trajectory and vice versa.
We use 20 [mm] in our later experiment.
Note that safety is always guaranteed because the smoothed trajectory is verified by a geometric collision checker at the end of our pipeline.

\subsection{Integration of CFN into a motion planning pipeline} \label{subsec:algorithm_performance}

\begin{figure}[b]
  \vspace{-8mm}
  \begin{minipage}[b]{1.0\hsize}
    \centering
    \includegraphics[height=5cm]{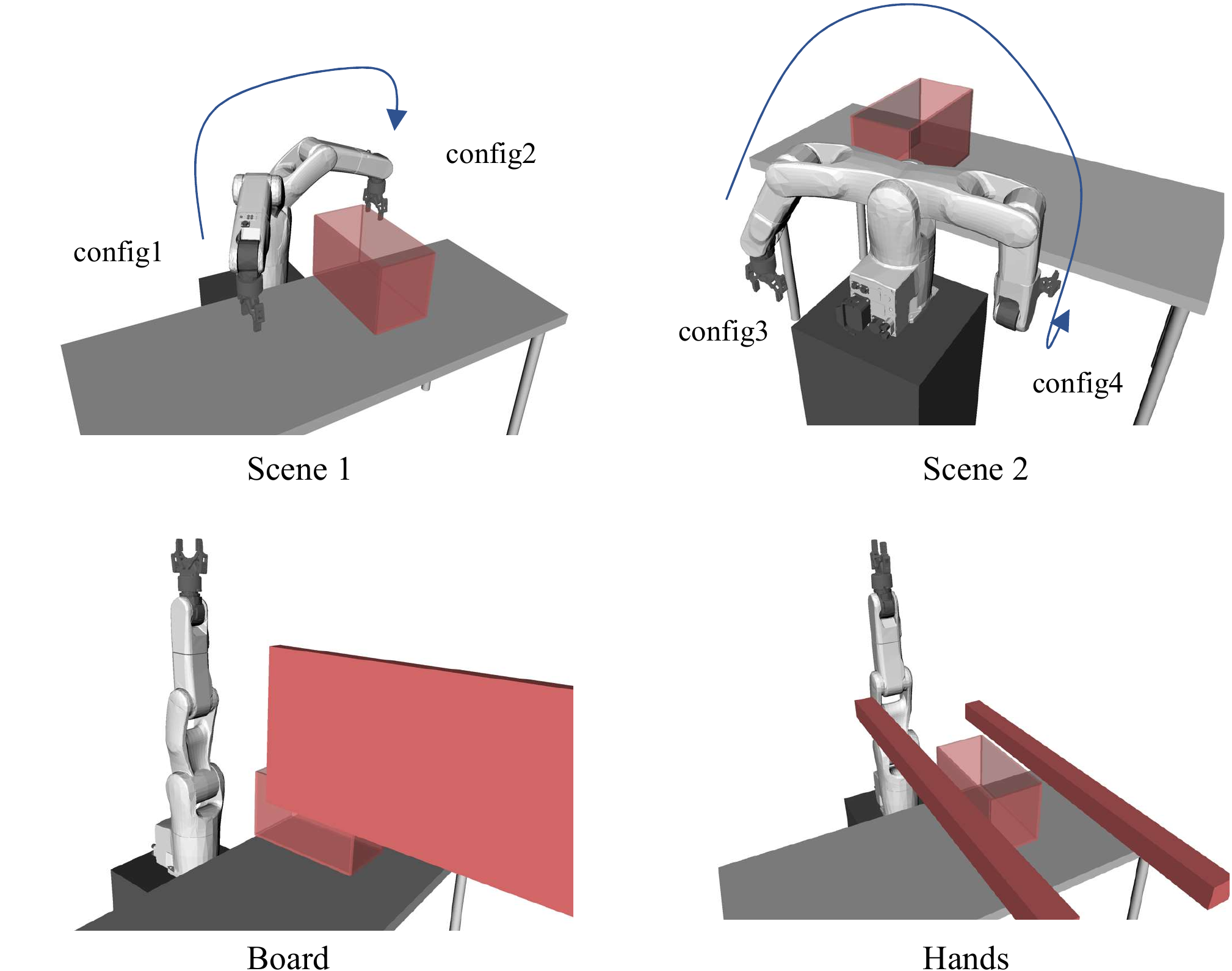}
    \label{fig:denso_scene}
  \end{minipage}
  \caption{Scenes and Obstacles for trajectory smoothing computation time comparison experiments}
  \label{fig:scenes}
\end{figure}

\begin{figure*}[thbp]
  \begin{center}
  \includegraphics[width=1.0\linewidth,trim={2.8cm 0.8cm 1.3cm 3.4cm},clip]{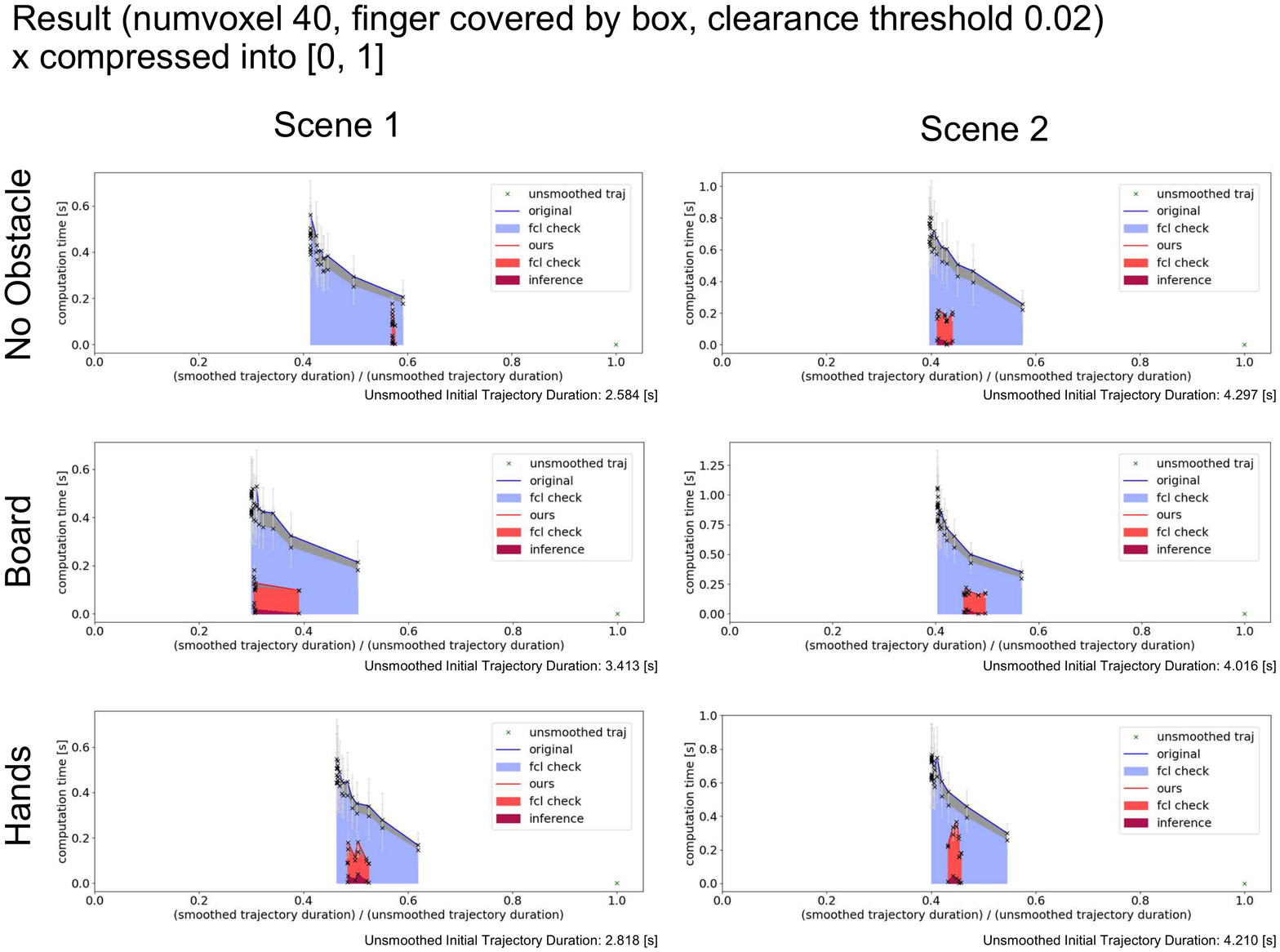}
  \caption{Reduction Ratio of Smoothed Trajectory Duration to Unsmoothed Trajectory Duration to Shortcutting
    Computation Time [s]. `original' is OpenRAVE's state-of-the-art parabolic smoother
    \cite{openrave_parabolicsmoother, hauserFastSmoothingManipulator2010}.  `fcl check' represents
    geometric collision checking by FCL\cite{FCL} and `inference'
    represents collision inference time.
    See \cref{subsec:algorithm_performance} for detail.
    }
  \label{fig:shortcut_trajtime_comparison}
  \end{center}
  \vspace{-4mm}
\end{figure*}

Secondly, we compare our proposed smoother using the trained neural
network with OpenRAVE's state-of-the-art parabolic smoother
\cite{openrave_parabolicsmoother} which originates from \cite{hauserFastSmoothingManipulator2010} and is updated with recent theoretical advancement\cite{lertkultanonTimeoptimalParabolicInterpolation2016}.
In this experiment, given a piecewise linear trajectory from a start to a goal
(from `config1' to `config2' in `Scene1', and from `config3' to
`config4' in `Scene2' shown in \cref{fig:scenes}), smoothers smooth
the trajectory and we measure their computation time and the duration
of its smoothed trajectory under different configurations, changing the maximum number of
iterations for the existing method, and the number of sampled waypoints
for our proposed method.  The clearance threshold for our
collision estimation is set to 20 [mm].  We sub-sampled shortcuts at
0.04 seconds interval.  Our code is based on
OpenRAVE\cite{diankov_thesis} and we use FCL\cite{FCL} as a collision
checking library.  All the experiments are done on a single machine,
on which Intel\textsuperscript{\textregistered}
Xeon\textsuperscript{\textregistered} W-2145 and GeForce GTX 1080 Ti are
mounted for CPU and GPU.

In \cref{fig:shortcut_trajtime_comparison}, we plot ratio of smoothed trajectory
duration to unsmoothed initial trajectory duration v.s. computation time for each method, where a shorter
trajectory and a faster computation time (the left, bottom side of the
figure) is preferable.  In the existing method, the smoothing time
increases as we increase the number of max iterations generating
shorter trajectory (blue line),
whereas in our method, the smoothing time does not
increase constantly as we increase the number of sampled configurations (red line).
The result shows that the computation speed of our proposed
method is generally 2--3x faster than that of the existing smoother to
generate trajectories of the same length, and can generate a much
shorter trajectory within the same computation time.

In some cases, e.g. Scene2 with Hands in
\cref{fig:shortcut_trajtime_comparison}, the time of geometric
collision checking increases to find another collision-free trajectory
when false negative collision detection happens, i.e., it can not
correctly estimate collision-free trajectories.  Statistically, the
first inferred collision-free candidate is actually collision-free in
86.1\% of 36 cases, and the second, third, and fourth candidates are
selected in 2.7\% (once), 5.6\% (twice) and 2.7\% (once) of the cases
consecutively.

In terms of optimality, our smoother cannot find an optimal trajectory as we see in \cref{fig:shortcut_trajtime_comparison}.
There are several reasons: First, we do not have multi-staged waypoint
sampling as it is done in \cite{hauserFastSmoothingManipulator2010},
therefore the robot does not accelerate well along the smoothed trajectory.
Second, we convert a pointcloud into voxels which include original obstacles.
This conversion adds voxel-size padding at maximum to obstacles.
Third, we set a clearance threshold in collision estimation, which can also be interpreted as padding and makes the robot larger than its real size, leading to a longer trajectory.
However, our focus is not on asymptotical performance but on reducing one-shot smoothing time against computation time, and this non-optimality is acceptable for our use case, i.e., realtime motion planning.

\begin{figure}[t]
  \begin{minipage}[b]{0.49\linewidth}
    \includegraphics[width=0.99\hsize]{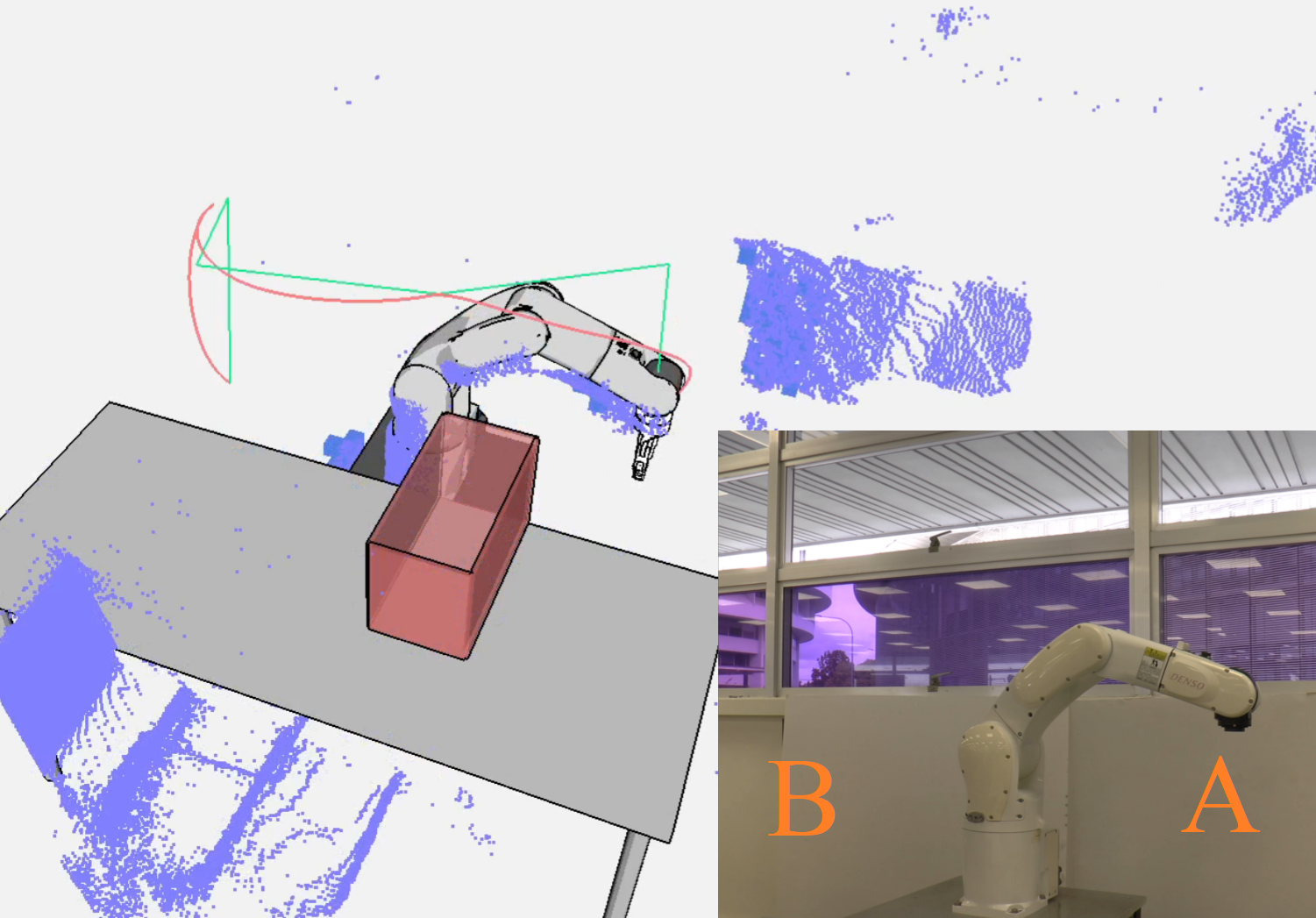}
    \subcaption{}
    \label{fig:realrobotexperiment_clip1}
  \end{minipage}
  \begin{minipage}[b]{0.49\linewidth}
    \centering
    \includegraphics[width=0.99\hsize]{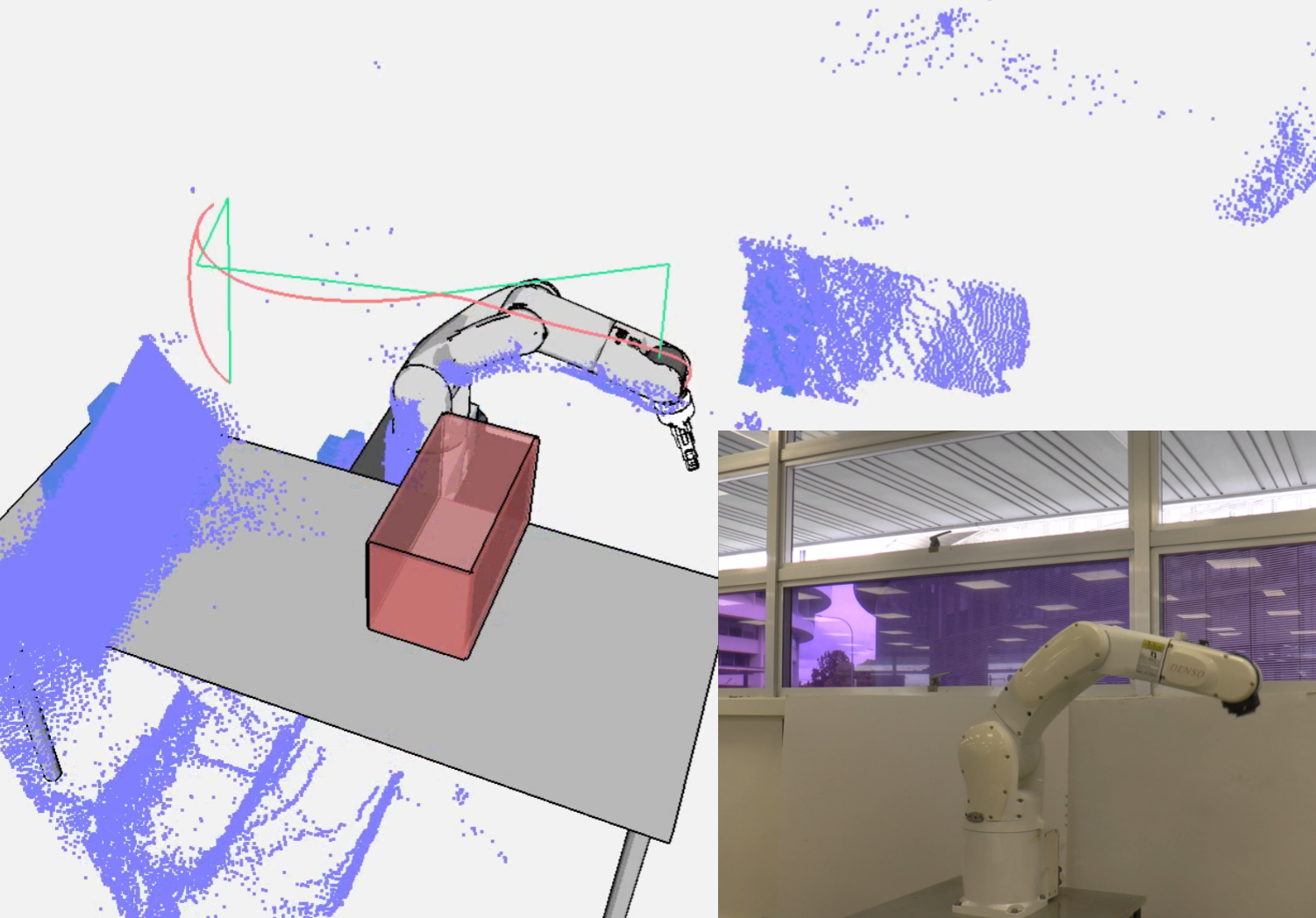}
    \subcaption{}
    \label{fig:realrobotexperiment_clip2}
  \end{minipage}
  \begin{minipage}[b]{0.49\linewidth}
    \centering
    \includegraphics[width=0.99\hsize]{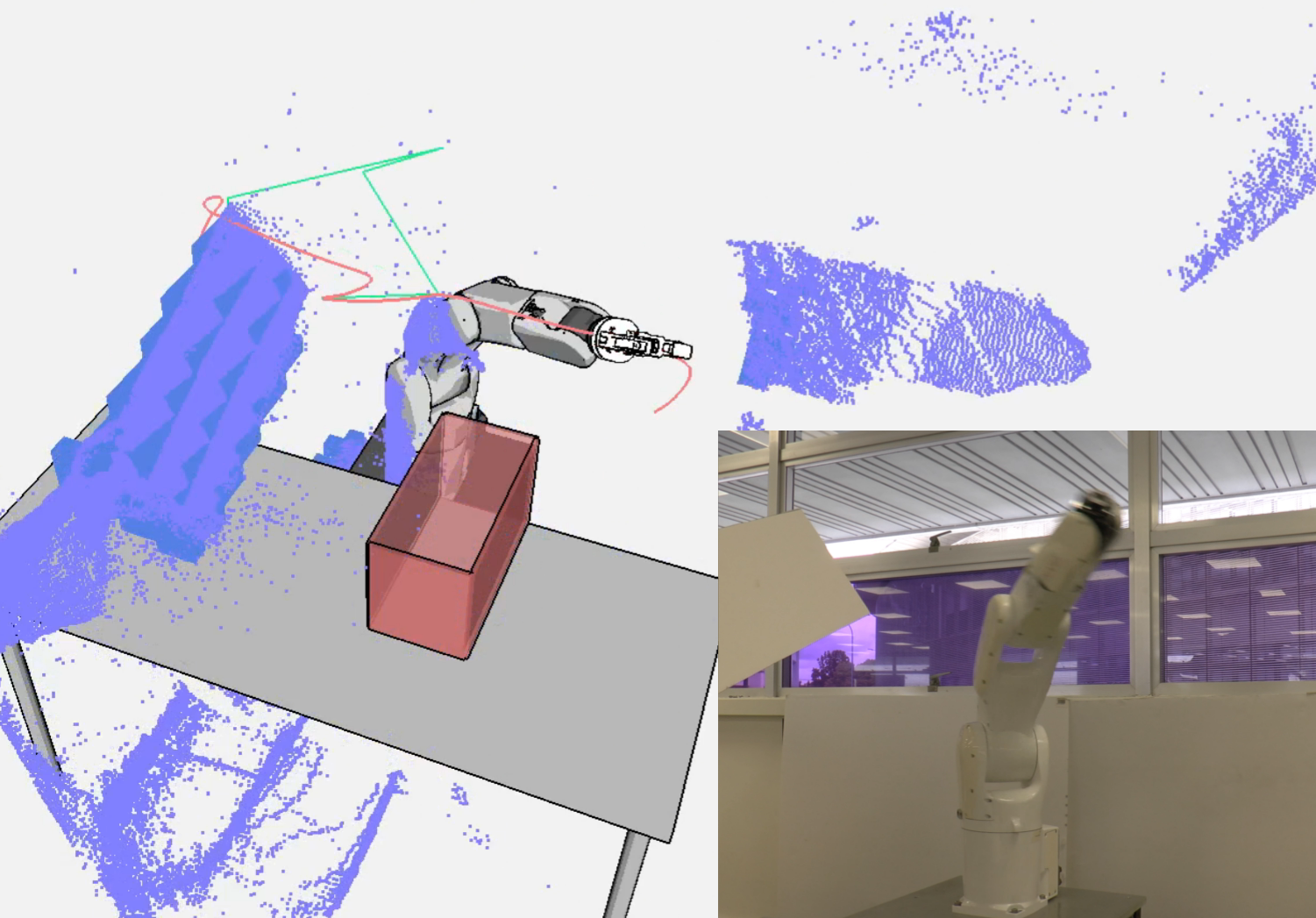}
    \subcaption{}
    \label{fig:realrobotexperiment_clip3}
  \end{minipage}
  \begin{minipage}[b]{0.49\linewidth}
    \includegraphics[width=0.99\hsize]{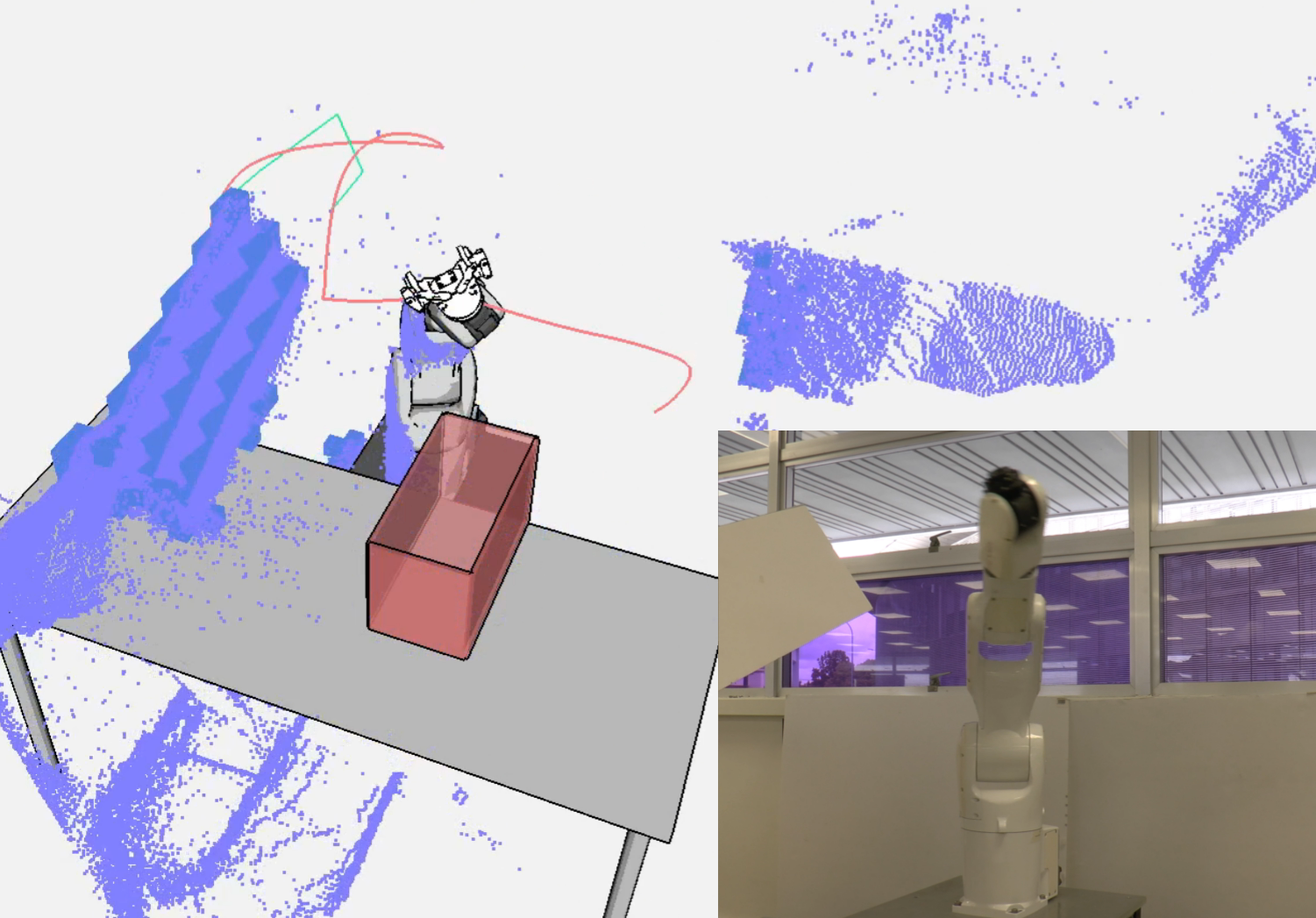}
    \subcaption{}
    \label{fig:realrobotexperiment_clip4}
  \end{minipage}
  \begin{minipage}[b]{0.49\linewidth}
    \centering
    \includegraphics[width=0.99\hsize]{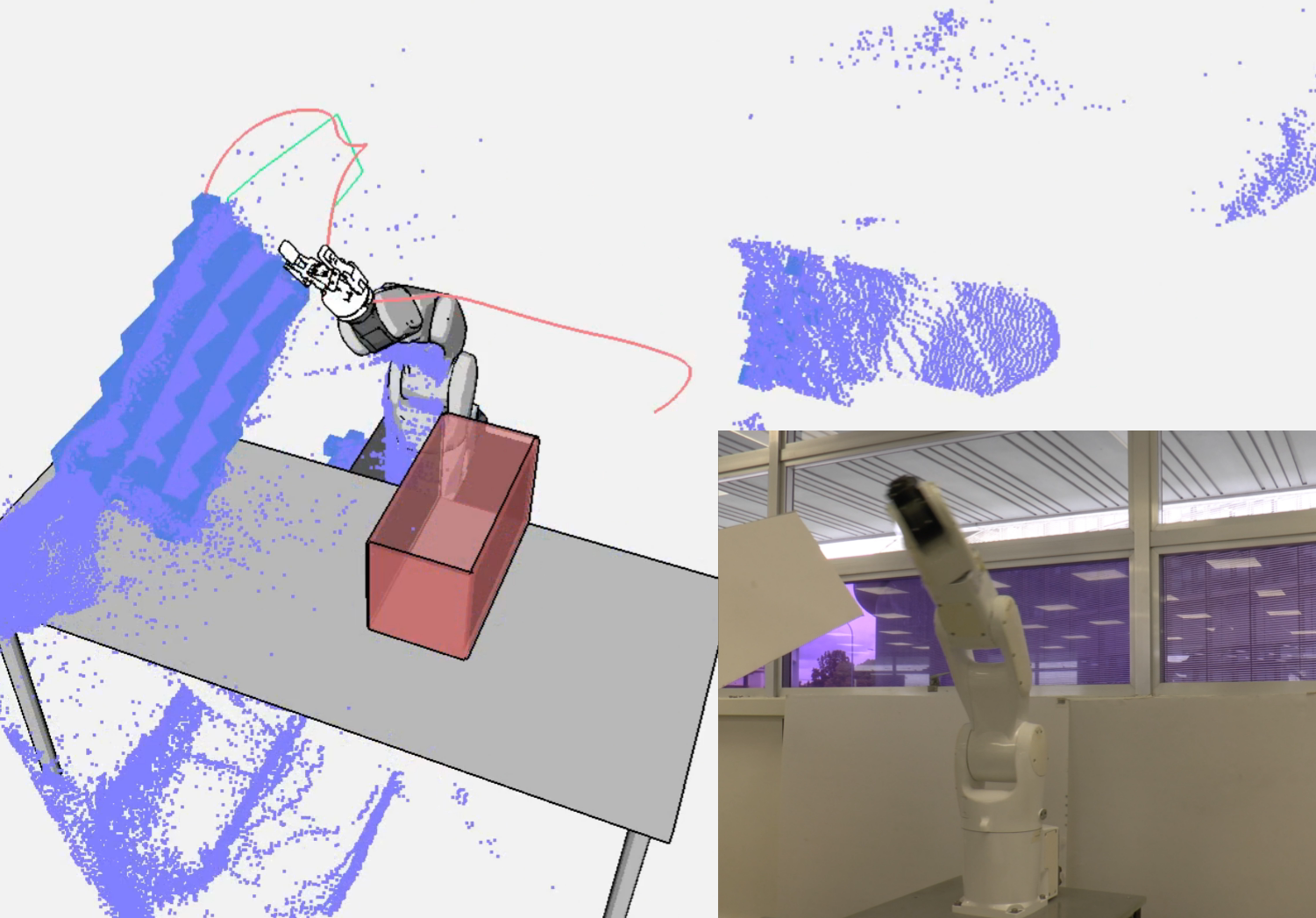}
    \subcaption{}
    \label{fig:realrobotexperiment_clip5}
  \end{minipage}
  \begin{minipage}[b]{0.49\linewidth}
    \centering
    \includegraphics[width=0.99\hsize]{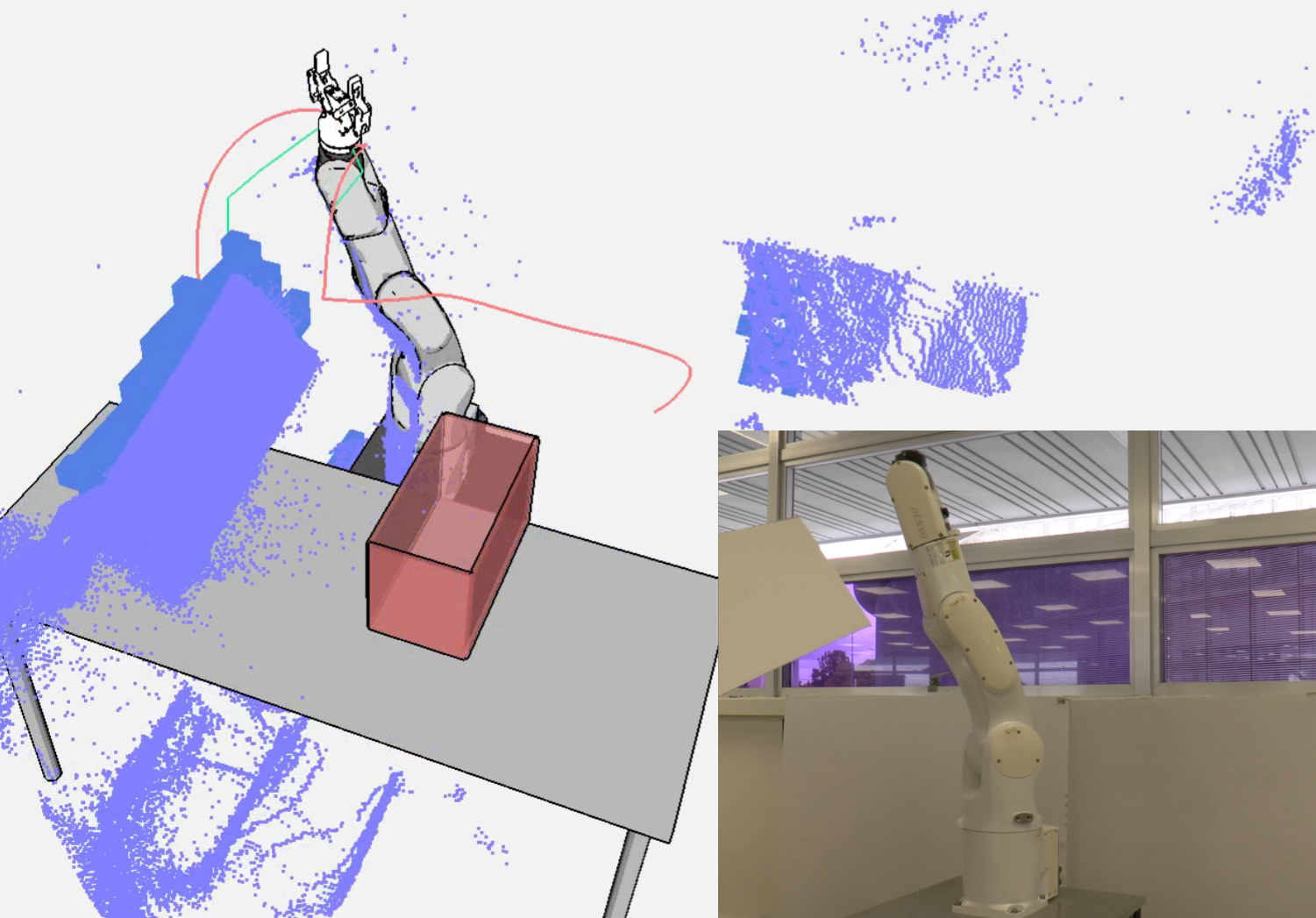}
    \subcaption{}
    \label{fig:realrobotexperiment_clip6}
  \end{minipage}
  \caption{Realtime Motion Planning and Smoothing on a physical
    robot. The robot loops between point A and point B while the
    experimenter randomly introduces an obstacle on the robot path.
    Purple points: raw point cloud from Kinect v2. Blue boxes: occupied
    voxels obtained by filtering the raw point cloud. Green line: piecewise
    linear trajectory output by realtime PRM. Pink line: trajectory
    smoothed in real time by our smoother. (a) Initial trajectory is
    planned and the robot starts moving. (b) The experimenter
    introduces the obstacle \emph{after} the robot has started
    moving. (c, d, e, f) As the obstacle approaches and collides with
    the planned trajectory, replanning with smoothing is triggered and
    the robot smoothly avoids the obstacle. The full experiment can be
    viewed at \url{https://youtu.be/XQFEmFyUaj8}.}
  \label{fig:realrobotexperimentclips}
  \vspace{-6mm}
\end{figure}

Finally, we conducted a physical robot experiment
(\cref{fig:realrobotexperimentclips}). The robot loops between point A
and point B while the experimenter randomly introduces obstacles on
the robot path.
The obstacles are monitored by a Kinect v2 mounted at the side of the workspace, and the obtained pointcloud is converted into a Voxel Occupancy Vector by thresholding the number of points in every voxel at a certain threshold (50 in our experiment) to reduce the effect of sensor noise.
By bringing the computation time of the entire
Vision--Planning--Execution loop under 300ms, our smoother enables the
robot to react to fast perturbations (we observed that using
OpenRAVE's state-of-the-art parabolic smoother, such realtime reaction
was not possible as the robot had to stop, take time to compute a new
trajectory, and restart), while being smooth (compare with Realtime
Robotics' demo \url{https://vimeo.com/359773568}).


\section{Conclusion}
\label{sec:conclusion}

We have proposed a trajectory smoother by leveraging a neural network
to estimate clearances and collisions between a robot and voxels in a
parallelized manner. Our planner is 2--3x faster than an existing
method, making realtime performance possible.

Why is our smoother faster and can generate shorter trajectory within
the same amount of computation time?  The reason for shorter
trajectory is that our smoother aggressively tries to connect distant
waypoints.  Even when the number of sampled waypoints is small, our
smoother tries to connect waypoints far away from each other (one of
them connects a starting point to a goal), so that the computed
trajectory tends to be shorter.  In contrast, with the existing
method, when the number of iterations is small, there is less
possibility to connect distant waypoints to have a shorter shortcut.
The reason why it is faster is because NN collision estimator can
\emph{batch-evaluate} many waypoints using significantly less time
than geometric collision checker.

Currently, there are a number of limitations, which we intend to
address in future work.

\begin{itemize}

\item Memory consumption and scalability: To find an optimal trajectory as
much as possible, we need to increase the number of sampling and have
a smaller voxel size, which leads to a larger batch size of input and
large memory consumption.
Under the configuration in \cref{sec:experiment}, our neural network model itself takes about 1.5 GB of GPU memory,
and we allocate 0.8 GB of GPU memory for temporary variables. 
This memory consumption increases in $O(c^2 V L)$ where $c$ is the number of sampled waypoints, $V$ is the number of voxels and $L$ is the length of the trajectory ($c^2$ because we have $K=\frac{(c+2)(c+1)}{2}$ shortcut candidates).
Therefore we need to prepare a GPU with large memory to increase the number of sampling waypoints, decrease the resolution of voxels, and apply for a large environment.

\item Planning Constraints: Our smoother assumes that collision checking
is the bottleneck of trajectory smoothing, i.e., the first
shortcut-candidates computation step is much faster than collision
checking.  If we need to take into account other constraints such as
torque limits and robot hand/grasped object's orientation, the first
shortcut computation may become a bottleneck of a whole smoothing
pipeline, and this will consequently decrease the speed of our
trajectory smoothing.  In this case, we need ways to
estimate torque and orientation in parallel to utilize our smoothing
method.

\end{itemize}


\bibliographystyle{unsrt}
\bibliography{sample}

\end{document}